\documentclass[letterpaper, 10 pt, conference]{ieeeconf}  

\IEEEoverridecommandlockouts                              

\usepackage{cite}
\usepackage{multicol}
\usepackage{amsmath,amssymb,amsfonts}
\usepackage{graphicx}
\usepackage{textcomp}
\usepackage{subcaption}
\usepackage{xcolor}
\usepackage{ragged2e}
\usepackage{booktabs}
\usepackage{algpseudocode}
\usepackage[ruled,linesnumbered, vlined]{algorithm2e}
\DontPrintSemicolon
\SetCommentSty{myCommentStyle}

\usepackage{amsfonts}
\let\labelindent\relax
\usepackage{enumitem}
\usepackage{amsmath}
\usepackage{graphicx}
\usepackage{subfig}
\usepackage[margin=0.792in]{geometry}
\usepackage{bbm}
\usepackage{algpseudocode} 
\usepackage{multirow}
\usepackage{xcolor}
\algnewcommand{\Inputs}[1]{%
  \State \textbf{Inputs:}
  \Statex \hspace*{\algorithmicindent}\parbox[t]{.8\linewidth}{\raggedright #1}
}
\algnewcommand{\Initialize}[1]{%
  \State \textbf{Initialize:}
  \Statex \hspace*{\algorithmicindent}\parbox[t]{.8\linewidth}{\raggedright #1}
}

\def\BibTeX{{\rm B\kern-.05em{\sc i\kern-.025em b}\kern-.08em
    T\kern-.1667em\lower.7ex\hbox{E}\kern-.125emX}}

\title{\LARGE \bf
Multi-Agent Monte Carlo Tree Search for Makespan-Efficient Object Rearrangement in Cluttered Spaces
}

\author{Hanwen Ren$^+$, Junyong Kim$^+$, Aathman Tharmasanthiran and Ahmed H. Qureshi
\thanks{* $^+$ denotes equal contribiton.}
\thanks{Hanwen Ren, Junyong Kim, Aathman Tharmasanthiran and Ahmed H. Qureshi are with the Department of Computer Science, Purdue University, West Lafayette, IN, USA, 47907. Email {\tt\small$\{$ren221, kim3722, atharmas, ahqureshi$\}@$purdue.edu}}%
}

\begin{document}

\thispagestyle{empty}
\pagestyle{empty}


\twocolumn[{
\begin{@twocolumnfalse}
\maketitle
\begin{center}
    \captionsetup{type=figure}
    \includegraphics[trim={0.4cm 0cm 0.4cm 0cm},clip,width=\textwidth]{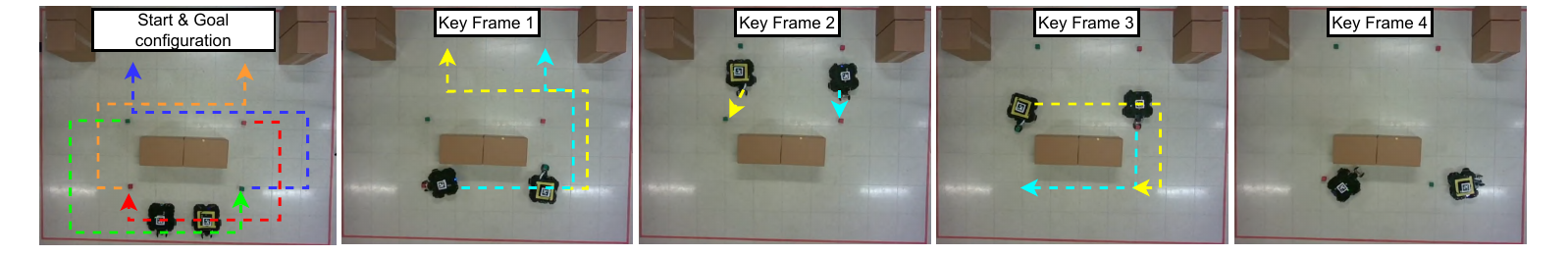}
    \captionof{figure}{The figure illustrates a non-monotone four-object relocation task. The left image shows the objects' start and goal states, while the others depict key pick-and-place robot actions, leading to the desired object rearrangement configuration. }
    \label{fig:real_exp_2}%
\end{center}
\end{@twocolumnfalse}
}]
\begingroup
\renewcommand{\thefootnote}{}
\footnotetext{$^+$ denotes equal contribution.}
\footnotetext{Hanwen Ren, Junyong Kim, Aathman Tharmasanthiran and Ahmed H. Qureshi are with the Department of Computer Science, Purdue University, West Lafayette, IN, USA, 47907.}
\endgroup



\begin{abstract}
Object rearrangement planning in complex, cluttered environments is a common challenge in warehouses, households, and rescue sites. Prior studies largely address monotone instances, whereas real-world tasks are often non-monotone—objects block one another and must be temporarily relocated to intermediate positions before reaching their final goals. In such settings, effective multi-agent collaboration can substantially reduce the time required to complete tasks. This paper introduces Centralized, Asynchronous, Multi-agent Monte Carlo Tree Search (CAM-MCTS), a novel framework for general-purpose makespan-efficient object rearrangement planning in challenging environments. CAM-MCTS combines centralized task assignment—where agents remain aware of each other’s intended actions to facilitate globally optimized planning—with an asynchronous task execution strategy that enables agents to take on new tasks at appropriate time steps, rather than waiting for others, guided by a one-step look-ahead cost estimate. This design minimizes idle time, prevents unnecessary synchronization delays, and enhances overall system efficiency. We evaluate CAM-MCTS across a diverse set of monotone and non-monotone tasks in cluttered environments, demonstrating consistent reductions in makespan compared to strong baselines. Finally, we validate our approach on a real-world multi-agent system under different configurations, further confirming its effectiveness and robustness.
\end{abstract}

\section{Introduction}
Object rearrangement planning is a common task in daily life—for instance, organizing shelf items for efficient space usage or sorting newly delivered packages in logistics centers. They enable warehouses to automate stock management and reduce operational costs, and in disaster sites, allow rescue teams to deploy robots to clear blocked paths by moving debris. Multi-agent systems are especially suited for these tasks, as collaboration among robots shortens the makespan and improves efficiency compared to a single robot. An example is illustrated in Fig. \ref{fig:real_exp_2} where two robots work together to solve a non-mononte relocation task.

Although object rearrangement appears natural to humans, it poses significant challenges for multi-agent systems. First, the problem is generally NP-hard \cite{reif1994motion, wilfong1988motion} even in the single-robot setting, since planning must account for both high-level task allocation and low-level motion generation. Second, additional factors further increase complexity: in non-monotone instances, some objects must be moved multiple times if their target region is initially occupied, and environmental constraints such as collision avoidance add further difficulty. Finally, the multi-agent setting compounds these challenges—task planners must evaluate numerous task assignment combinations for high-quality solutions, while motion planners must generate collision-free trajectories for all agents in constrained workspaces.

Most existing work addresses object rearrangement planning with a single robot \cite{wang2022efficient}, typically using fixed-arm manipulators for tabletop rearrangement or mobile robots in confined environments \cite{10801607, kim2025integrating}. These approaches generally rely on tree search, where nodes represent environment states connected by single-robot actions, expanding from an initial configuration until all objects are relocated. Recent efforts have extended the problem to multi-agent systems, but most decentralized methods \cite{ma2017lifelong, 9410352} are limited to monotone pickup-and-placement tasks. Approaches that address non-monotone rearrangements \cite{10298262} are typically learning-based, requiring large amounts of demonstration data and thus incurring high training costs. Moreover, they enforce synchronous execution: all agents must complete their current tasks before new ones are assigned. Consequently, even robots that finish early are forced to wait, leading to unnecessarily long makespans during inference.


This paper introduces the Centralized, Asynchronous, Multi-agent Monte Carlo Tree Search (CAM-MCTS) approach for monotone and non-monotone object rearrangement planning in complex, cluttered environments. Our method centrally plans for the entire multi-agent system while enabling agents to act asynchronously—a feature typically associated with decentralized methods. This asynchronous behavior allows agents to take on new tasks without waiting for others to finish, fully exploiting multi-agent collaboration and yielding makespan-efficient solutions. In summary, the main contributions of this paper are:
\begin{itemize}[leftmargin=*]
\item A centralized task assignment module that allocates tasks based on each agent’s current status—whether occupied with a previously assigned task or idle—significantly reducing the search space.
\item An asynchronous task execution strategy that enables agents to take on new tasks at appropriate time steps—rather than waiting for others—using a one-step look-ahead cost estimate.
\item A centralized, asynchronous multi-agent planner that leverages the MCTS paradigm to efficiently generate object rearrangement solutions with short makespans.
\end{itemize}
We evaluate our approach on a wide range of monotone and non-monotone rearrangement planning tasks through comprehensive tests in both simulation and real-world settings. The results show consistent improvements in task completion success rate and reduced makespan compared to strong baselines.
\section{Related Work}
The most relevant work to CAM-MCTS includes methods on multi-agent pickup and delivery, object rearrangement planning, and Monte Carlo Tree Search. We discuss these studies below and explicitly highlight how our approach differs from and advances beyond prior work.

The \textbf{Multi-Agent Pickup and Delivery (MAPD)} problem \cite{zong2022mapdp} seeks task and motion plans for a multi-agent system to complete a set of pickup and delivery operations. For example, \cite{9410352} introduces marginal cost-based and regret-based task selection strategies to assign the most suitable tasks to agents and minimize total travel delay. Similarly, \cite{liu2019task} formulates task assignment as a Traveling Salesman Problem (TSP) \cite{bektas2006multiple}, generating task sequences for each agent via cost estimation to reduce makespan. Decoupled MAPD techniques have also been developed for lifelong online settings, such as the token-passing with task swaps method \cite{ma2017lifelong}, where agents sequentially plan paths and can even reassign tasks that have not yet been executed. 

However, these MAPD approaches inherently assume monotone rearrangement settings, where start and goal regions remain feasible throughout execution \cite{salzman2020research,xu2022multi}. As a result, they are not suitable for real-world non-monotone tasks, where objects often block one another and must be relocated to intermediate positions. In contrast, CAM-MCTS explicitly addresses both monotone and non-monotone object rearrangement, enabling effective planning in complex, maze-like environments.


\textbf{Object rearrangement planning} \cite{guo2023recent} is a well-studied problem in robotics, particularly within Task and Motion Planning (TAMP) \cite{garrett2021integrated, srivastava2014combined}. The problem is generally NP-hard \cite{reif1994motion, wilfong1988motion}, as it requires jointly reasoning over high-level task planning and low-level motion planning. Most existing studies address the problem in various environments using a single-agent setting \cite{ 9036915, 10610092}. Compared to a single agent, multi-agent systems can solve complex tasks more efficiently \cite{8352646}. However, applying multi-agent systems introduces significant challenges: the task assignment space grows combinatorially \cite{6386013}, and the motion planning problem becomes more complex. 

To address these challenges, a multi-agent learning-based framework was proposed in \cite{10298262}, which iteratively selects objects, determines relocation regions, and assigns them to robots. However, it relies on large amounts of demonstration data, leading to high training costs, and assumes synchronized task execution, forcing all robots to wait for others before proceeding. In contrast, our approach is non-learning-based and enables asynchronous execution, allowing agents to take on new tasks as soon as they are available, thereby reducing idle time and improving overall efficiency.

\textbf{Monte Carlo Tree Search (MCTS)} is a general algorithm that combines the precision of tree search and the generality of random sampling \cite{ chaslot2008monte}. Leveraging the Monte Carlo simulation process, MCTS requires little or no domain knowledge; thus, it achieves tremendous success in artificial intelligence in general \cite{browne2012survey}. In the game of Go, AlphaGo and AlphaZero \cite{silver2018general, moerland2018a0c} combine MCTS with a trained neural network to predict the search probabilities in the next move and the game-winner in the current stage, beating the highest-ranked human go player at the moment. Driven by the success of MCTS in game AI, it has also been applied to other domains such as scheduling, planning, and combinatorial optimization \cite{swiechowski2023monte}. For example, authors of \cite{10801607} address the non-monotone object rearrangement planning problems in narrowly confined environments through a decoupled multi-stage MCTS. The work \cite{eiffert2020path} merges MCTS with generative Recurrent Neural Networks (RNN) for path planning in dynamic environments and demonstrates decent performance through improved motion prediction accuracy. Although MCTS is used mainly as a centralized method to generate synchronous solutions, our method introduces the asynchronous feature, leading to lower makespan.

\section{Problem Formulation}
Let a workspace be denoted as $\mathcal{W} \subseteq \mathbb{R}^2$. Within the space, there exist a set of non-movable obstacles $\mathcal{B}$, a group of objects $\mathcal{O} = \{o_1,\dots,o_n\}$ and a team of robot agents $\mathcal{Q} = \{q_1,\dots, q_m\}$. The obstacles $\mathcal{B}$ form environmental boundaries that prevent the agents from navigating through. At any time step $t$, the locations of each agent $p_q^t$ form the agent location set $\mathcal{P}_Q^t \subseteq \mathcal{W}\backslash \mathcal{B}$ while the locations of each object $p_o^t$ form the object location set $\mathcal{P}_O^t \subseteq \mathcal{W}\backslash \mathcal{B}$. Each agent has the ability to pick up an object and place it in a feasible location. We denote the agent $q$'s action of picking up object $o_i$ as $r_{(q, o_i)}^{pick}$ while the action of placing object $o_i$ as $r_{(q, o_i)}^{place}$. The path of the agent associated with each pick or place action is represented as an ordered sequence of agent locations $v_{(q,o_i)}^{pick/place} = [p_{q}^t, p_{q}^{t+1},\dots, p_{q}^{t+k}]$. Note that in the above-mentioned agent path, the agent location can remain the same for a certain number of time steps, meaning it can choose to stay idle. Then, the task assignment of an individual agent $q_i$ can be described as a sequence of pick-and-place actions $\pi_{q} = \{(r_{(q, o_i)}^{pick}, r_{(q, o_i)}^{place}),\dots, (r_{(q, o_j)}^{pick}, r_{(q, o_j)}^{place})\}$ through path $v_{q} = \{(v_{(q, o_i)}^{pick}, v_{(q, o_i)}^{place}),\dots, (v_{(q, o_j)}^{pick}, v_{(q, o_j)}^{place})\}$. Finally, the collection of all agents' task assignments forms the global task assignment for the multi-agent system $\pi = \{\pi_{q_i},\dots,\pi_{q_m}\}$ with the corresponding global path $V = \{v_{q_i},\dots, v_{q_m}\}$. Finally, for a certain global task assignment $\pi$, under the assumption that the agent takes one time step to move to an adjacent location, the makespan $T$ is defined as the maximum time steps needed across all agents, which is also equal to the maximum path length, thus $T = \max_{k=1}^m \lvert v_{q_k} \rvert$.

In object arrangement planning tasks, the inputs are composed of an object start location configuration $a^s = \{p_{o_1}^s,\dots,p_{o_n}^s\}$, an object goal location configuration $a^g = \{p_{o_1}^g,\dots,p_{o_n}^g\}$ and an agent starting location configuration $\{p_{q_1}^s,\dots, p_{q_m}^s\}$. Depending on whether some objects need to be relocated to a buffer region before the actual goal, the tasks can be further classified into monotone and non-monotone instances. Given the inputs, our aim is to find a collision-free global task assignment for the multi-agent system $\pi=\{\pi_{q_1},\dots, \pi_{q_m}\}$ with associated path $V = \{v_{q_1},\dots, v_{q_m}\}$ that relocates the objects from the start configuration $a^s$ to the goal configuration $a^g$ with minimal makespan $T$ for both monotone and non-monotone instances. The problem can be formally written as finding the global policy $\pi$ with associated path $V$ such that
\begin{multline}
    \text{minimize} \quad  \max\nolimits_{k=1}^m \lvert v_{q_k}\rvert \text{    subject to} \tag{1} \\ 
  \pi(a^s) = a^g  \text{ and }p^t_{q_k} \neq p^t_{q_u}~~~\forall~k \neq u, ~\forall~ 0 \leq t \leq T 
\end{multline} 

\section{Method}

This section introduces our novel Centralized, Asynchronous, Multi-Agent Monte Carlo Tree Search (CAM-MCTS) approach for object rearrangement planning in complex and cluttered environments. Existing literature \cite{swiechowski2023monte} claims that MCTS is perfectly suitable for solving problems that can be modeled as a Markov Decision Process (MDP) \cite{puterman1990markov}. Thus, in our centralized search tree, each node stores all information about the current problem-solving process, including the current object configuration $a^t = \{p_{o_1}^t,\dots, p_{o_n}^t\}$, the goal object configuration $a^g = \{p_{o_1}^g,\dots, p_{o_n}^g\}$, and the current status for each agent $q_i$. The agent's status comprises its current location $p_{q_i}^t$ and the object in hand $o_j$. In addition, parent and child tree nodes are linked by asynchronous task execution with the corresponding agents' path $v = \{v_{q_1},\dots,v_{q_m}\}$. The following paragraphs disclose the details of our CAM-MCTS by fitting them into the standard MCTS paradigm composed of selection, expansion, simulation, and back-propagation.

\subsection{Selection}
In each iteration, CAM-MCTS selects a leaf node by navigating through successive child nodes with the maximum value of the Upper Confidence Bound (UCB) \cite{garivier2011upper}. In our design, the UCB is constructed as follows
\begin{equation}\tag{2}\label{ucb}
    UCB = -\frac{\sum_{i=1}^h(\sum_{k=1}^m \vert v_{q_k}\rvert + \alpha\max_{k=1}^m \lvert v_{q_k}\rvert)}{B_{n}} +2\sqrt{\frac{\log B_{p}}{B_{n}}} 
\end{equation}
The numerator of the first term is the negation of the accumulative cost which is defined as a weighted combination between the multi-agent system's total traveling distance $\sum_{k=1}^m \lvert v_{q_k}\rvert$ and the makespan $\max_{k=1}^m \lvert v_{q_k}\rvert$ across all the $h$ simulations performed so far. The tunable hyperparameter $\alpha$ balances the cost between the moving distance and the makespan. By maximizing the UCB value during the selection process, we choose the most promising leaf nodes with small moving distances and short makespan. The $B_p$ and $B_n$ in Equation \ref{ucb} are the visiting counts for the parent and selected nodes, respectively.

\subsection{Expansion}
Based on the status of the selected node, the expansion module proposes various subsequent task and motion plans to bring it one step closer to solving the problem. The two main components of our design are the centralized task assignment approach and the asynchronous task execution strategy.
\subsubsection{Centralized Task Assignment}
The centralized task assignment method creates multiple potential plans and then assigns the most suitable tasks to each agent for individual plans. At any time step during the object rearrangement planning process, all agents with objects in hand are grouped into the active agent set, while the remaining are labeled as the idle set. Depending on the groups that the agent belongs to, different strategies are applied.

The active agents $Q_{active}$ are forced to keep relocating their objects in hand. However, the relocation region is determined based on the categories that each active agent falls into. The first type of agents are those that just arrives at the pickup region of the object $o_j$, it needs to verify whether the goal location of the object $a^g[o_j]$ is occupied. The goal location is labeled as occupied if there is another object that no agent has picked up. If the goal region is free, the desired relocation region is set directly to the goal. Otherwise, a buffer location $p^b_{o_j}$ is purposed as a temporary destination. Our approach proposes a buffer by sampling a position from a standard normal distribution centered at the goal. The buffer region is ensured to be feasible and not the current goal of other active agents. Sampling buffer location near the goal aims to make optimistic progress in finishing the task as the final move from the buffer region to the actual goal is minimized. When multiple buffer locations are generated simultaneously, they are forced to be mutually distinguishable. The second type of active agent are those during their journey for object relocation to the buffer, as proposed earlier. Our method re-verifies the feasibility of the goal locations of the objects in hand. If the goal region is free, the agents change their destinations from the buffers to the actual goals. Otherwise, the agents continue with their current path to the buffer. Finally, the tasks of the third type of agents on the way to the goal region are unchanged.

On the other hand, multiple potential task assignment plans exist for idle agents $Q_{idle}$. Our approach creates all combinations of the remaining objects with the number of the idle agents. Then, for each combination, the agents are set to pick up the nearest object based on the Euclidean distance, which is an admissible heuristic used in various search-based methods for cost estimation \cite{liu2011comparative}. The final task assignments for the multi-agent system are generated by combining each of the idle agent plans with the active agent plan.
\subsubsection{Asynchronous Task Execution}
\begin{figure}[t]
\includegraphics[trim={0cm 0.8cm 0cm 0.5cm},clip, width=8cm]{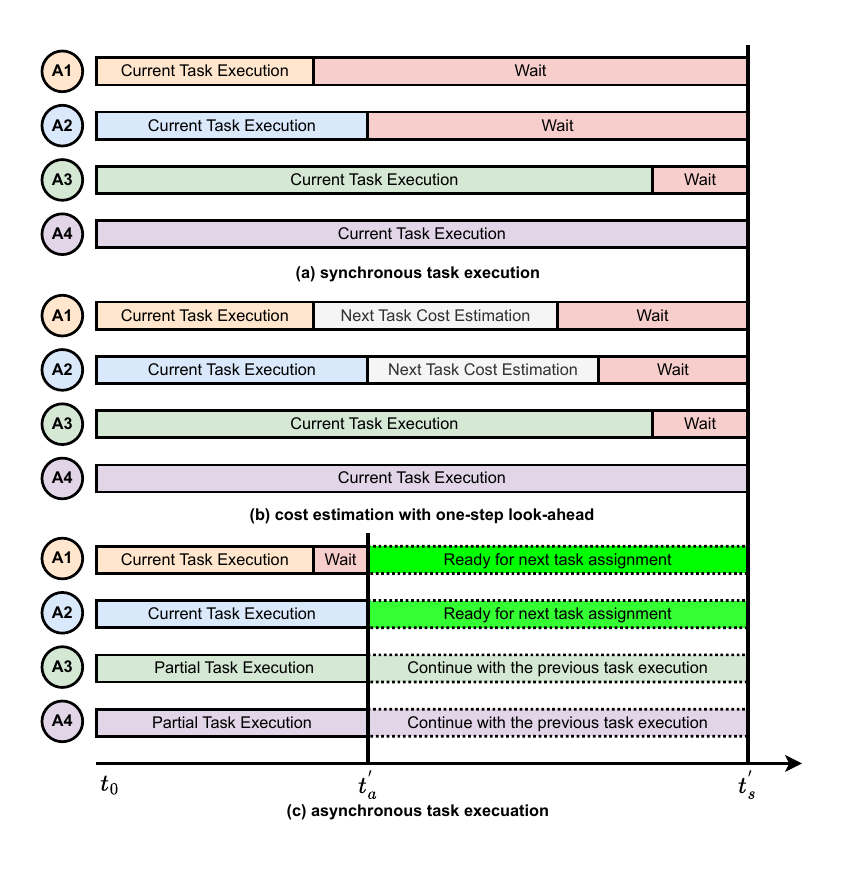}
\centering
\caption{The figure illustrates the asynchronous task execution design for agent A1 to A4. Starting from the synchronous task execution (a), our method leverges the cost estimation with one-step look-ahead (b) to find the most appropriate time step $t'_a$ (c) to terminate the current iteration. In this way, agent A1 and A2 can start performing new tasks without waiting for A3 and A4.}%
    \label{fig: ate}\vspace{-0.3in}
\end{figure}

Following the task assignments, our task execution module computes the associated path and then proposes the most appropriate time step to terminate the current iteration so that agents who complete their tasks earlier can start taking new ones, making task execution asynchronous. The main advantage of the asynchronous design over the synchronous one is the increased makespan efficiency, as the agents are not forced to wait for the last one to finish before taking on the next available task. Thus, we propose a novel asynchronous task execution design that leverages the cost estimation with one-step look-ahead. The high level idea is illustrated in Fig. \ref{fig: ate}. In the example shown in the figure, our method ends the current iteration early as waiting for Agent 3 prevents Agent 1 and 2 from completing the next available task. Our approach balances the solution quality and the scalability, resulting in makespan-efficient solutions for long-horizon object rearrangement planning tasks. 

\begin{algorithm}[hbt!]
\caption{Asynchronous Task Execution}\label{alg:ate}
\KwData{$\pi = \{p_{q_1},\dots,p_{q_m}\}$}
$v = \text{ICBS}(\pi)$ \Comment{synchronous task execution path}\; 
$\{(t_i, q_i)\}_{\{m\}} = \text{FindFinishTime}(v).\text{sort}()$\;
$t_h = \infty$ \Comment{tolerance horizon}\;
$t' = \text{None}$ \Comment{asynchronous execution end time}\;
\For{$(t_i, q_i) \in \{(t_i, q_i)\}_{\{m\}}$}{
    \If{$t_i \leq t_h$}{
         $t'$ = $t_i$\;
         \If{$q_i \in Q_{idle}$}{
            $t_h = \min(t_h, t_i + \min_{j=1}^k \text{C2G}(p_{q_i}, p_{o_j}))$
         }
         \ElseIf{$q_i \in Q_{active}$ }{
            $t_h = \min(t_h, t_i + \text{C2G}(p_{q_i}, p_{o_j}))$
         }
    }
    \Else{
        break\;
    }
}
\For {$v_{q_i} = [p_{q_i}^t,\dots, p_{q_i}^{t+k}] \in v$}{
    $v_{q_i} = v_{q_i}[0:t']$ \;
}
\Return $v$ \Comment{asynchronous task execution path}\;
\end{algorithm}
\vspace{-0.0in}
Algorithm \ref{alg:ate} depicts the workflow of the algorithm. First, given a task assignment $\pi$, we first leverage the improved conflict-based search (ICBS) approach \cite{boyarski2015icbs} to generate the initial synchronous task execution path $v$ (Line 1). ICBS creates a path for each agent in a decentralized manner and grows a binary constraint tree to gradually resolve the conflicts. Then, given the initial path $v$, our method sorts the agents in increasing order based on their task completion time (Line 2). In the next step, we iterate through the sorted agent list to check whether its waiting time is long enough to take the next available task by a one-step look-ahead mechanism. If so, we want to end the execution early to make the waiting agents available for new tasks. To achieve this design, we set a tolerance horizon $t_h$, indicating the maximum time step during which all waiting agents cannot possibly finish any future tasks. Then, for each agent in the sorted list, if its task completion task $t_i$ is earlier than the tolerance horizon $t_h$, we set the asynchronous execution end time $t'$ to $t_i$ and update the tolerance horizon $t_h$ based on the following two cases. First, suppose the agent becomes idle at the completion time $t_i$. In that case, its associated tolerance horizon is the time step to get to the start location of the nearest object approximated by the Cost-to-Go (C2G) function (Lines 8-9). However, if the agent has an in-hand object at time $t_i$, the tolerance horizon is set to be the predicted time to rearrange the in-hand object to the goal location (Lines 10-11). The cost-to-go function predicts the task completion time by multiplying the Manhattan distance between the start and goal regions with a cost ratio. This cost ratio tracks the average time step to move one unit in the workspace based on all the generated paths in the search tree. On the other hand, the agents with task completion time $t_i$ later than $t_h$ will not finish their task in the current iteration as waiting for them prevents other agents from completing the next potential task (Lines 12-13). Finally, the agents' paths are segmented based on $t'$, making task execution asynchronous (Lines 14-15).

Leveraging the current status of the problem and the two previously introduced submodules, the MCTS expansion method creates a fixed number of new tree nodes based on possible task assignments and then associates each of them with the corresponding asynchronous task execution path.

\subsection{Parallel simulation and Sequential Back-propagation}
The simulation process iteratively grows a pathological tree from an unvisited tree node until the problem is solved. In our CAM-MCTS, the simulation step for unvisited tree nodes is performed in parallel for better time efficiency. Recall the UCB calculation shown in Equation \ref{ucb}. For unvisited tree nodes, their UCB value will be infinity as the visited count $B_n$ is 0, meaning a simulation process must be carried out for all expanded tree nodes. Due to the MDP nature of the search, performing simulation in parallel for all the newly created tree nodes yields the same result as doing it in sequence. However, when simulating all nodes in parallel, rewards are cached instead of back-propagated and used only when the corresponding node is later selected.
\subsection{Result Extraction}
Our CAM-MCTS declares success when the resulting tree node in the selection process returns one in which all the tasks are fulfilled, i.e., all the objects have been relocated from the start configuration to the goal configuration. The final asynchronous execution path $V = \{v_{q_1},\dots, v_{q_m}\}$ and the global policy $\pi$ can be recovered by a backward tree traversal until the root followed by a reverse operation. 

\begin{table*}[!ht]
  \fontsize{9}{10}\selectfont
  \begin{center}
    \begin{tabular}{c |c c c c | c c c c |c c c c}
      \toprule
      \multirow{2}{*}{\textbf{(\#object, \#agent)}}& \multicolumn{4}{c|}{\textbf{CAM-MCTS (Ours)}} & 
      \multicolumn{4}{c|}{\textbf{RSP}} &
      \multicolumn{4}{c}{\textbf{CAM-UCS}}\\
       & SR $\uparrow$ & PT $\downarrow$ & TD $\downarrow$ & MS  $\downarrow$ & SR $\uparrow$ & PT $\downarrow$ & TD $\downarrow$ & MS  $\downarrow$ & SR $\uparrow$ & PT $\downarrow$ & TD $\downarrow$ & MS  $\downarrow$ \\
       \midrule
       (6,2) & \textbf{100.0} & 2.7 & 430 & 239 & 83.3 & \textbf{0.4} & 479 & 329 & 58.3 & 160.8 & \textbf{385} & \textbf{205}\\
       (6,3) & \textbf{100.0} & 5.8 & 405 & 163 & 100.0 & \textbf{0.4} & 435 & 247 & 62.5 & 115.7 & \textbf{372} & \textbf{145}\\
       (7,2) & \textbf{100.0} & 5.6 & \textbf{512} & \textbf{281} & 100.0 & \textbf{0.7} & 574 & 426 & 0.0 & - & - & -\\
       (7,3) & \textbf{100.0} & 6.5 & 502 & 202 & 100.0 & \textbf{0.5} & 555 & 323 & 8.3 & 220.2 & \textbf{397} & \textbf{150}\\
       (8,2) & \textbf{100.0} & 10.2 & \textbf{562} & \textbf{306} & 83.3 & \textbf{0.7} & 650 & 448 & 0.0 & - & - & -\\
       (8,3) & \textbf{100.0} & 12.7 & 575 & 227 & 100.0 & \textbf{0.5} & 613 & 349 & 4.2 & 1.8 & \textbf{393} & \textbf{172}\\
        (8,4) & \textbf{100.0} & 12.6 & 532 & 168 & 95.8 & \textbf{0.6} & 547 & 244 & 12.5 & 115.7 & \textbf{437} & \textbf{133}\\
       (9,2) & \textbf{100.0} & 13.4 & \textbf{616} & \textbf{337} & 95.8 & \textbf{0.8} & 669 & 467 & 0.0 & - & - & -\\
       (9,3) & \textbf{100.0} & 31.2 & \textbf{608} & \textbf{235} & 100.0 & \textbf{0.6} & 655 & 362 & 0.0 & - & - & -\\
       (9,4) & \textbf{100.0} & 23.5 & \textbf{600} & \textbf{185} & 100.0 & \textbf{0.6} & 614 & 293 & 0.0 & - & - & -\\
       (10,2) & \textbf{100.0} & 21.6 & \textbf{733} & \textbf{400} & 91.7 & \textbf{1.0} & 799 & 553 & 0.0 & - & - & -\\
       (10,3) & \textbf{100.0} & 39.7 & \textbf{720} & \textbf{275} & 100.0 & \textbf{1.0} & 774 & 445 & 0.0 & - & - & -\\
       (10,4) & \textbf{100.0} & 40.9 & \textbf{720} & \textbf{219} & 100.0 & \textbf{0.9} & 771 & 362 & 0.0 & - & - & -\\
      \bottomrule
    \end{tabular}
    \caption{This table compares our CAM-MCTS approach against baseline methods. Overall, CAM-MCTS achieves the best performance in terms of success rate, solution makespan, and scalability. While CAM-UCS performs well in scenarios with fewer objects, it suffers from very high planning times (PT) and its success rate drops to zero as task difficulty increases, demonstrating poor scalability. RSP achieves lower PT and comparable success rates to ours but yields extremely high makespans, limiting its practical utility. In conclusion, CAM-MCTS delivers the best overall performance: it scales to complex scenarios, maintains the lowest makespan, and achieves tractable planning times, making it an ideal planner for multi-agent rearrangement tasks.}
    \label{tab:object RP result}
  \end{center}
  \vspace{-0.2in}
\end{table*}

\begin{figure}[t]
\centering
\begin{subfigure}{0.23\textwidth}
\centering
\includegraphics[trim={0cm 0cm 0cm  0cm},clip, width=0.95\linewidth]{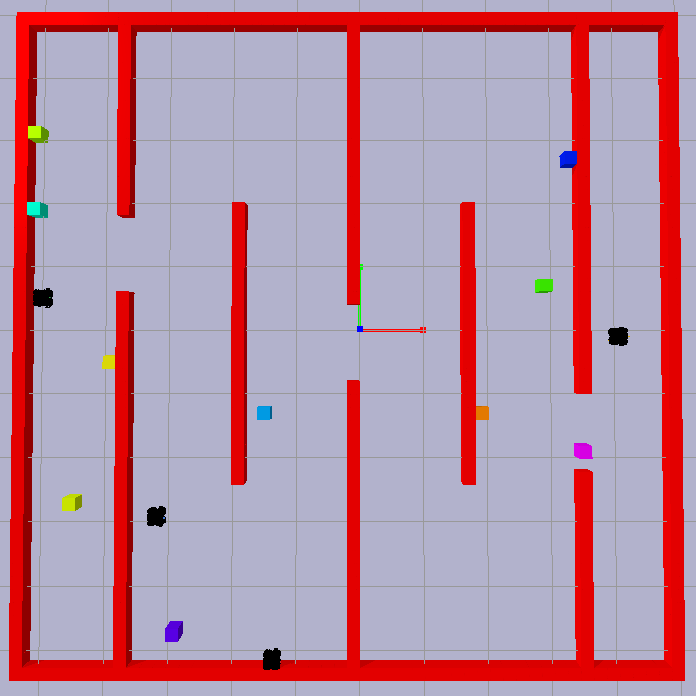}
\caption{narrow passage}
\end{subfigure}
\begin{subfigure}{0.23\textwidth}
\centering
\includegraphics[trim={0cm 0cm 0cm 0cm},clip,width=0.95\linewidth]{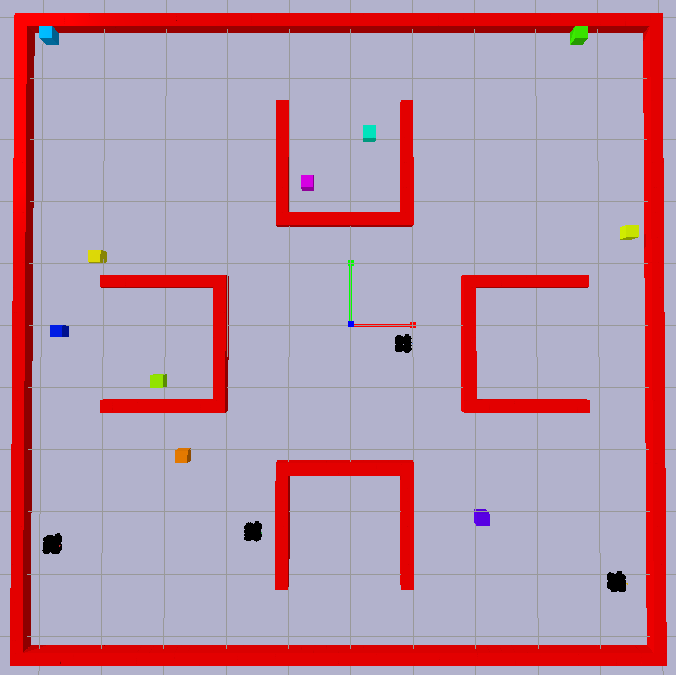}
\caption{warehouse}
\end{subfigure}
\begin{subfigure}{0.23\textwidth}
\centering
\includegraphics[trim={0cm 0cm 0cm 0cm},clip,width=0.95\linewidth]{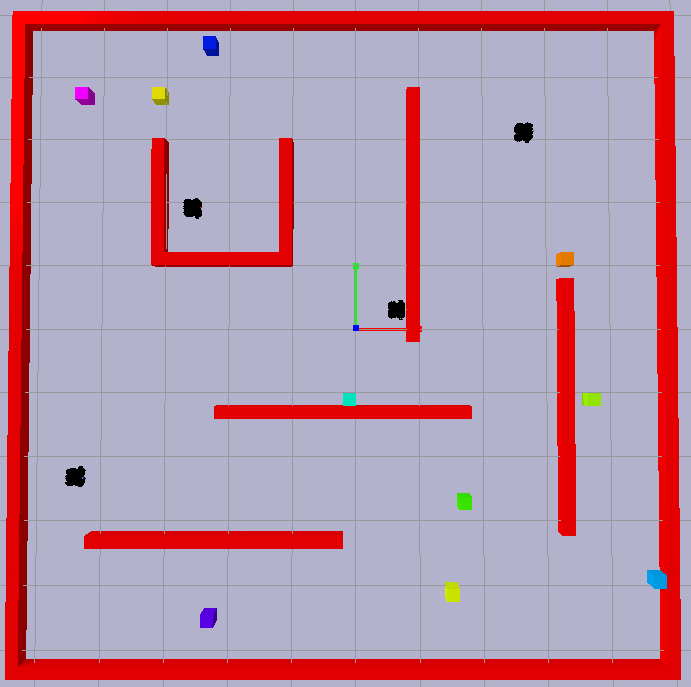}
\caption{random obstacles}
\end{subfigure}
\begin{subfigure}{0.23\textwidth}
\centering
\includegraphics[trim={0cm 0cm 0cm 0cm},clip,width=0.95\linewidth]{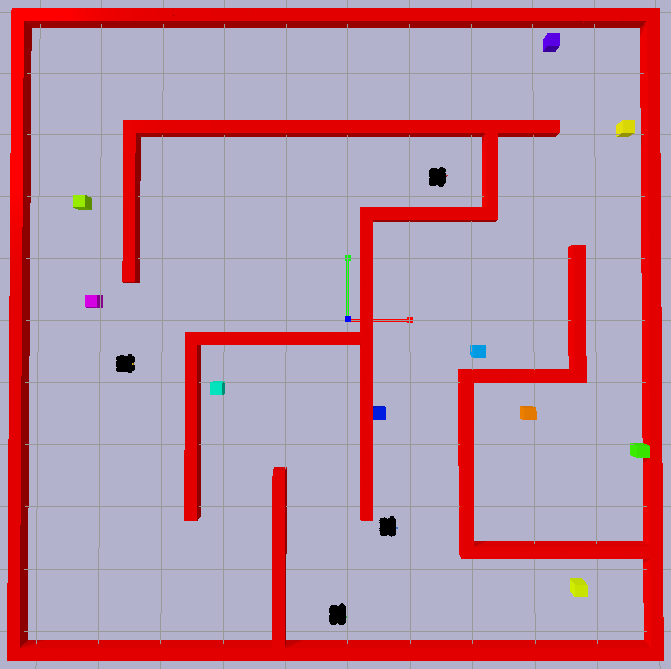}
\caption{maze}
\end{subfigure}
\caption{This figure depicts the narrow passage, the warehouse, the random obstacles and the maze environments. The colored cubes are the objects while the block ones are the robots.}
\label{fig:simulation}
\vspace{-0.2in}
\end{figure}

\begin{figure}[t]
\includegraphics[trim={3cm 0cm 3cm 2.3cm},clip, width=0.48\textwidth]{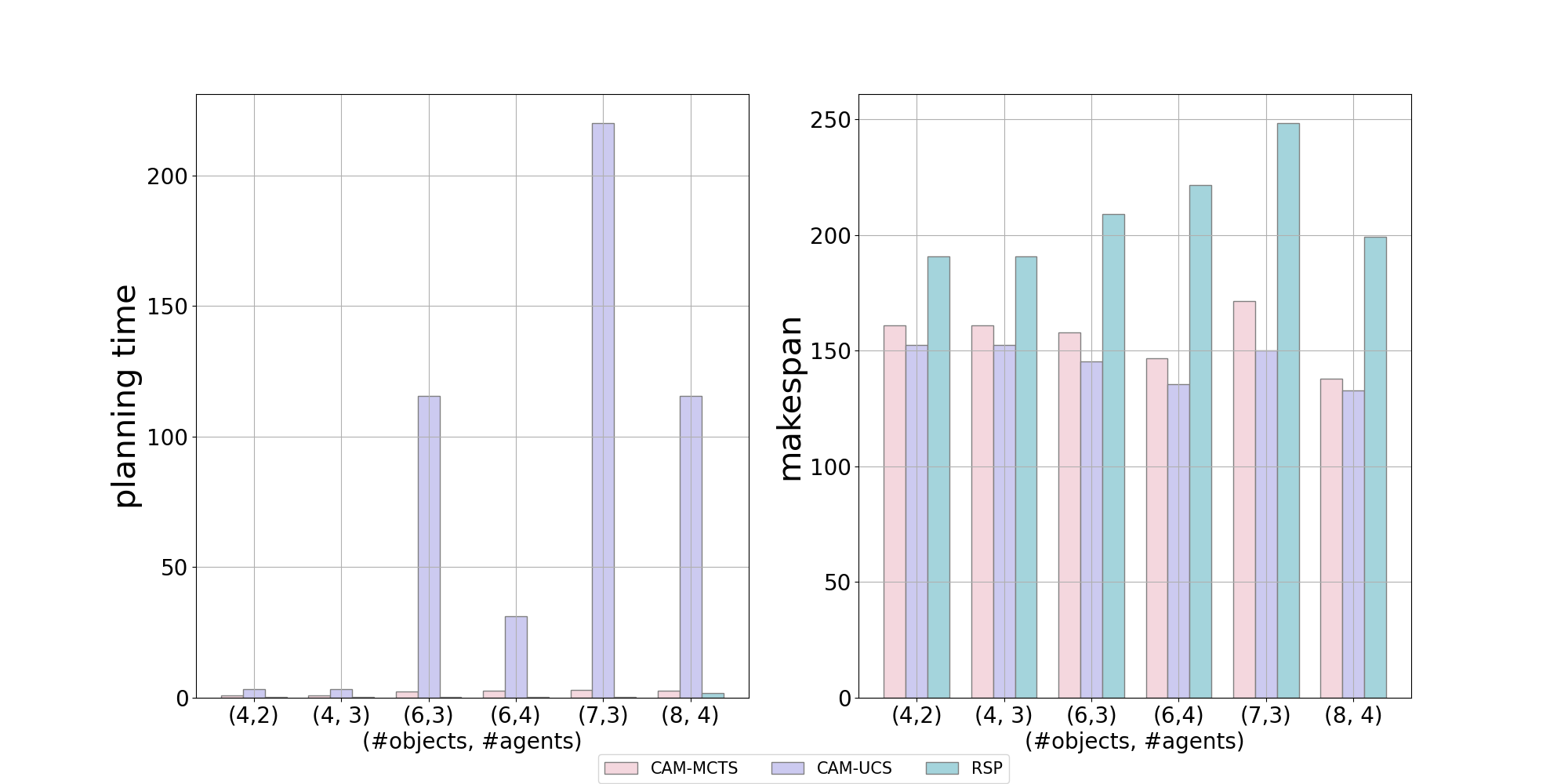}
\centering
\caption{These results compare CAM-MCTS, CAM-UCS, and RSP only in scenarios where all methods succeed, and should be interpreted in conjunction with Table 1. As shown, CAM-UCS achieves slightly better makespan in tasks with fewer objects, but only by a narrow margin over CAM-MCTS—and at the cost of significantly higher planning times. RSP, on the other hand, achieves low planning times but suffers from the highest makespan. Taken together, these results highlight that CAM-MCTS provides the best overall trade-off between planning time, makespan, and scalability to more challenging scenarios.}%
    \label{fig: sim plot}\vspace{-0.2in}
\end{figure}

\section{Experiments}
In this section, we present the simulation setup, the baseline comparison, the ablation studies, and the real-world experiments. 
\subsection{Simulation Setup}
We create four complex simulation environments representing the narrow passage, the warehouse, the random obstacles, and the maze setting, each with a size of (50$\times$50). For each scene, we put 4-10 objects for a team consisting of 2-4 agents for relocation, creating 20 object-agent combinations ranging from (2 agents, 4 objects) to (4 agents, 10 objects). We include four agents in scenarios with more than seven objects, as task allocation would otherwise be trivial. Then, for each combination, the objects' start and goal configurations are set to follow three schemes: (1) Random - The start and goal are randomly generated. (2) Sorting - The start is randomly generated, while the goal follows a structured pattern. (3) Shuffling - The start and goal share the same set of locations but are arranged in different configurations. The random and sorting scenarios covers a mixture of monotone and non-monotone object rearrangement planning tasks, while the shuffling cases only comprise non-monotone ones. In total, we create 480 test instances for a comprehensive evaluation. Examples of testing instances in diverse environments are shown in Fig. \ref{fig:simulation}. To quantitatively evaluate the performance, we use the following metrics
\begin{itemize}[leftmargin=*]
    \item \textbf{Success Rate (SR)}: Success rate is the percentage of solved cases, with failures defined as those not completed within 300 seconds.
    \item \textbf{Planning Time (PT)}: The planning time tracks the time consumption for the planners to solve the instance.  
    \item \textbf{Traveling Distance (TD)}: This metric records the total traveling distance among all agents in the object rearrangement planning process.
    \item \textbf{MakeSpan (MS)}: As described in the problem formulation section, the makespan quantifies the total time step needed for the rearrangement planning tasks. 
\end{itemize}

\begin{figure*}[t]
\centering
    \includegraphics[trim={0.6cm 0.5cm 0 0cm},clip,width=15cm]{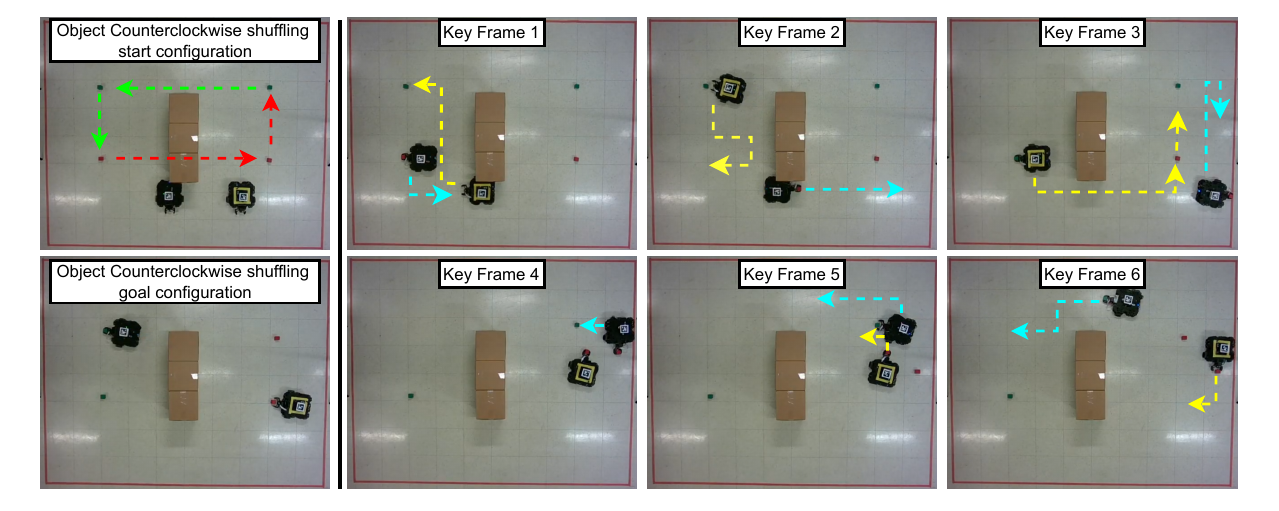}
    \caption{The figure shows a non-monotone four objects counterclockwise shuffling case. The start and goal configurations are shown on the left, with the red and green arrows indicating the goal location for each object. The rest of the images denote the keyframes involving object pick and place. Our CAM-MCTS finds the asynchronous solutio in 0.83 seconds, while the actual execution lasts 394 seconds.}%
    \label{fig:real_exp}%
    \vspace{-0.1in}
\end{figure*}

\subsection{Simulation Results}

\subsubsection{Baseline Comparison}
We consider the following two baselines to compare with our approach:
\begin{itemize}[leftmargin=*]
    \item \textbf{Random Synchronous Planner (RSP)}: RSP is a synchronous tree search method that keeps randomly creating a task assignment based on the current status of the problem until it is solved.
    \item \textbf{Centralized, Asynchronous, Multi-agent Uniform Cost Search (CAM-UCS):} In this approach, we substitute the MCTS framework with the uniform cost search approach \cite{felner2011position}. CAM-UCS uses the same cost function as CAM-MCTS, expanding the lowest-cost node and guaranteeing the shortest makespan.
\end{itemize}
Note that prior rearrangement planning approaches primarily focus on monotone tasks and are therefore inapplicable to our scenarios. The only recent learning-based multi-agent method for non-monotone cases is Maner \cite{10298262}, but it requires large amounts of training data. In contrast, our approach is non-learning-based and avoids this limitation. 

Table \ref{tab:object RP result} reports results across increasing difficulty levels, with all metrics averaged over successful instances except for success rate. Figure \ref{fig: sim plot} compares the makespan and planning time only in scenarios where all methods succeed.

RSP runs faster than CAM-MCTS but exhibits poor makespan. As task complexity grows, CAM-MCTS consistently outperforms RSP in success rate, travel distance, and makespan. This advantage stems from CAM-MCTS’s centralized task assignment module, which generates multiple high-quality candidate plans and evaluates them through MCTS simulations. By contrast, RSP randomly selects valid assignments, often discarding optimal choices. As tasks deepen, these random decisions accumulate, leading to instability and high variance in solution quality.

CAM-UCS achieves better makespan than CAM-MCTS in simple cases with few objects and agents, but its performance drops sharply beyond seven objects, with success rates falling to near zero due to time limit. UCS expands nodes with minimal current cost, ignoring long-term potential. While effective in small problems, this strategy forces UCS to explore far more states as complexity increases, severely reducing efficiency. MCTS, by contrast, expands the most promising nodes, improving overall scalability.

Finally, Fig. \ref{fig: sim plot} compares CAM-MCTS, CAM-UCS, and RSP in cases where all succeed (to be interpreted with Table~1). CAM-UCS shows slightly better makespan in small tasks, but only by a narrow margin and at much higher planning cost. RSP, while faster, suffers from the highest makespan. Overall, CAM-MCTS strikes the best balance between planning time, makespan, and scalability, making it the most effective planner in challenging multi-agent rearrangement tasks.

\begin{table*}[!ht]
  \fontsize{9}{10}\selectfont
  \begin{center}
    \begin{tabular}{c |c c c c | c c c c |c c c c}
      \toprule
      \multirow{2}{*}{\textbf{(\#object, \#agent)}}& \multicolumn{4}{c|}{\textbf{CAM-MCTS (Ours)}} & 
      \multicolumn{4}{c|}{\textbf{CAM-MCTS w/o look-ahead}} &
      \multicolumn{4}{c}{\textbf{CSM-MCTS}}\\
       & SR $\uparrow$ & PT $\downarrow$ & TD $\downarrow$ & MS $\downarrow$ & SR $\uparrow$ & PT $\downarrow$ & TD $\downarrow$ & MS $\downarrow$ & SR $\uparrow$ & PT $\downarrow$ & TD $\downarrow$ & MS  $\downarrow$ \\
       \midrule
       (6,2) & \textbf{100.0} & \textbf{2.7} & 430 & 239 & 100.0 & 7.4 & 438 & \textbf{233} & 100.0 & 2.9 & \textbf{420} & 270\\
       (6,3) & \textbf{100.0} & 5.8 & 405 & 163 & 100.0 & 15.8 & 422 & \textbf{156} & 100.0 & \textbf{2.9} & \textbf{383} & 192\\
       (7,2) & \textbf{100.0} & \textbf{5.6} & 512 & 281 & 100.0 & 17.8 & 508 & \textbf{268} & 91.7 & 7.5 & \textbf{484} & 317\\
       (7,3) & \textbf{100.0} & \textbf{6.5} & 502 & 202 & 100.0 & 31.9 & 518 & \textbf{196} & 100.0 & 9.8 & \textbf{477} & 245\\
       (8,2) & \textbf{100.0} & \textbf{10.2} & 562 & 306 & 100.0 & 21.6 & 580 & \textbf{301} & 95.8 & 11.6 & \textbf{540} & 341\\
       (8,3) & \textbf{100.0} & 12.7 & 575 & 227 & 100.0 & 61.5 & 590 & \textbf{217} & 95.8 & \textbf{11.2} & \textbf{538} & 270\\
       (8,4) & \textbf{100.0} & \textbf{12.6} & 532 & 168 & 95.8 & 41.6 & 550 & \textbf{161} & 91.7 & 21.7 & \textbf{508} & 196\\
       (9,2) & \textbf{100.0} & \textbf{13.4} & 616 & 337 & 100.0 & 26.7 & 610 & \textbf{317} & 95.8 & 15.5 & \textbf{568} & 371\\
       (9,3) & \textbf{100.0} & 31.2 & 608 & 235 & 100.0 & 114.8 & 617 & \textbf{228} & 91.7 & \textbf{20.3} & \textbf{565} & 265\\
       (9,4) & \textbf{100.0} & \textbf{23.5} & 600 & 185 & 100.0 & 81.7 & 613 & \textbf{178} & 100.0 & 24.1 & \textbf{563} & 225\\
       (10,2) & \textbf{100.0} & \textbf{21.6} & 733 & 400 & 100.0 & 56.7 & 742 & \textbf{389} & 70.8 & 36.1 & \textbf{686} & 438\\
       (10,3) & \textbf{100.0} & \textbf{39.7} & 720 & 275 & 91.6 & 129.6 & 740 & \textbf{260} & 91.7 & 48.0 & \textbf{669} & 321\\
       (10,4) & \textbf{100.0} & \textbf{40.9} & 720 & 219 & 79.2 & 139.0 & 754 & \textbf{218} & 91.7 & 55.0 & \textbf{691} & 278\\
      \bottomrule
    \end{tabular}
    \caption{The ablation study highlights the importance of both the one-step look-ahead and the asynchronous task execution strategy in addressing long-horizon multi-agent object rearrangement problems. CSM-MCTS relies on synchronous execution, where all agents wait for each other to finish before proceeding. Although this slightly reduces travel distance, it significantly increases the overall makespan. The one-step look-ahead strategy shows clear advantages. Although without it, the method achieved shorter makespans but at the cost of substantially higher planning times and lower success rates. Taken together, these results demonstrate that CAM-MCTS offers the best trade-off between planning time, travel distance, and makespan, establishing it as a versatile and effective planner for object rearrangement.}
    \label{tab:object RP result ablation}
  \end{center}
  \vspace{-0.3in}
\end{table*}

\subsubsection{Ablation Studies}
We also perform ablation studies to verify the effectiveness of our one-step look-ahead design in the asynchronous task execution module and the usage of the asynchronous framework with the following two methods:
\begin{itemize}[leftmargin=*]
    \item \textbf{CAM-MCTS w/o look-ahead:} This method replaces one-step look-ahead design in our method with the idea that terminating the iteration once an agent finishes, allowing immediate task reassignment.
    \item \textbf{Centralized, Synchronous, Multi-agent Monte Carlo Tree Search (CSM-MCTS)}: CSM-MCTS is the synchronous counterpart of our approach where all agents pick up and relocate simultaneously.
\end{itemize}

The results are summarized in Table \ref{tab:object RP result ablation}. Our method consistently outperforms the synchronous baseline CSM-MCTS in both success rate and makespan. This improvement stems largely from our asynchronous task execution strategy: instead of idling while other agents finish, agents immediately take on new tasks, boosting overall efficiency and reducing makespan. In addition, limiting each tree node to a fixed number of children prevents the search tree from growing uncontrollably, improving success rates on challenging instances. Compared to CAM-MCTS without look-ahead, our approach achieves higher success rates and better time efficiency in complex settings, though with slightly longer makespans. Overall, the combination of MCTS with asynchronous execution significantly enhances scalability for long-horizon, multi-agent rearrangement tasks, achieving a balanced trade-off among PT, TD, and MS.

\subsection{Real Robot Experiments}
We deploy our CAM-MCTS on a multi-agent system composed of two TurtleBot3 robots in an arena with dimensions of (300 cm, 300 cm). Five non-monotone rearrangement tasks involving four objects are constructed to verify the real-world performance of our approach, one example can be found in Fig. \ref{fig:real_exp_2}. During the experiment, our method succeeds in solving all cases with an average planning time of 0.45 seconds and an average execution time of 383 seconds. Fig. \ref{fig:real_exp} shows a successfully solved non-monotone object counterclockwise shuffling task. Our CAM-MCTS finds a valid plan in 0.83 seconds. In the generated plan, the robots perform asynchronous task execution (Key Frames 1-3) and collaborate with each other (Key Frames 4-6) to fulfill the object rearrangement task with a short makespan. The other scenarios are available in our supplementary videos.

\section{Conclusions \& Future Work}
This paper presents a novel centralized, asynchronous, multi-agent makespan-efficient planner for object rearrangement in complex, cluttered environments, addressing both monotone and non-monotone tasks. Our approach combines centralized task assignment with asynchronous task execution to generate diverse, high-quality task allocations for the multi-agent system. Leveraging the MCTS framework, the planner then produces efficient object rearrangement solutions with reduced makespan, significantly improving overall system performance. Experimental results show that our method consistently outperforms baseline approaches across varying difficulty levels in both monotone and non-monotone scenarios. For future work, we aim to further improve scalability to handle more complex cases and enhance the MCTS simulations by incorporating a learned reward prediction function for faster node evaluation.

\bibliographystyle{unsrt}
\bibliography{root}

\end{document}